\documentclass[letterpaper]{article} 
\usepackage{aaai2026}  
\usepackage{times}  
\usepackage{helvet}  
\usepackage{courier}  
\usepackage[hyphens]{url}  
\usepackage{graphicx} 
\usepackage{natbib}  
\usepackage{caption} 
\usepackage{algorithm}
\usepackage{algorithmic}
\usepackage{amsmath}
\usepackage{amssymb}
\usepackage{tabularray}
\usepackage{amsfonts}
\usepackage{xcolor}
\usepackage{appendix}
\usepackage{float}

\usepackage{newfloat}
\usepackage{listings}
\DeclareCaptionStyle{ruled}{labelfont=normalfont,labelsep=colon,strut=off} 
\floatstyle{ruled}
\newfloat{listing}{tb}{lst}{}
\floatname{listing}{Listing}
\title{MRT: Learning Compact Representations with Mixed RWKV-Transformer for Extreme Image Compression}
\author{
    Han Liu\textsuperscript{\rm 1} \quad
    Hengyu Man\textsuperscript{\rm 1,\thanks{Corresponding author}} \quad
    Xingtao Wang\textsuperscript{\rm 1,2}\quad
    Wenrui Li\textsuperscript{\rm 1}\quad
    Debin Zhao\textsuperscript{\rm 1}\quad
}
\affiliations{
    \textsuperscript{\rm 1}Harbin Institute of Technology \\
    \textsuperscript{\rm 2}Harbin Institute of Technology Suzhou Research Institute

    manhengyu@hotmail.com
}

\usepackage{bibentry}

\begin{document}

\maketitle

\begin{abstract}
    Recent advances in extreme image compression have revealed that mapping pixel data into highly compact latent representations can significantly improve coding efficiency. However, most existing methods compress images into 2-D latent spaces via convolutional neural networks (CNNs) or Swin Transformers, which tend to retain substantial spatial redundancy, thereby limiting overall compression performance. In this paper, we propose a novel Mixed RWKV-Transformer (MRT) architecture that encodes images into more compact 1-D latent representations by synergistically integrating the complementary strengths of linear-attention-based RWKV and self-attention-based Transformer models. Specifically, MRT partitions each image into fixed-size windows, utilizing RWKV modules to capture global dependencies across windows and Transformer blocks to model local redundancies within each window. The hierarchical attention mechanism enables more efficient and compact representation learning in the 1-D domain. To further enhance compression efficiency, we introduce a dedicated RWKV Compression Model (RCM) tailored to the structure characteristics of the intermediate 1-D latent features in MRT. Extensive experiments on standard image compression benchmarks validate the effectiveness of our approach. The proposed MRT framework consistently achieves superior reconstruction quality at bitrates below 0.02 bits per pixel (bpp). Quantitative results based on the DISTS metric show that MRT significantly outperforms the state-of-the-art 2-D architecture GLC, achieving bitrate savings of $43.75\%$, $30.59\%$ on the Kodak and CLIC2020 test datasets, respectively. All source code are
    available at \url{https://github.com/luke1453lh/MRT}.
\end{abstract}


\section{Introduction}
The explosive growth of visual data has intensified the demand for advanced image 
compression techniques capable of preserving high perceptual quality at extremely 
low bitrates. While traditional compression standards, e.g., H.266/VVC~\cite{VVC},
and learning-based neural image compression methods~\cite{groupedmixer, jiang2023mlic, Cheng_2020_CVPR, balle2018variational, 10155660,10510460} 
primarily optimize for distortion-oriented metrics such as PSNR and MS-SSIM, 
they often struggle to maintain semantic integrity and visual fidelity under 
severe bitrate constraints~\cite{blau2019rethinking}. To address these limitations, recent compression 
approaches based on generative models have emerged, leveraging powerful 
generative priors to enable high-fidelity reconstruction even in 
extreme-low-bitrate scenarios~\cite{finetuneVQGAN,illm,hific}.
\begin{figure}[t]
\centering
\includegraphics[width=0.9\columnwidth]{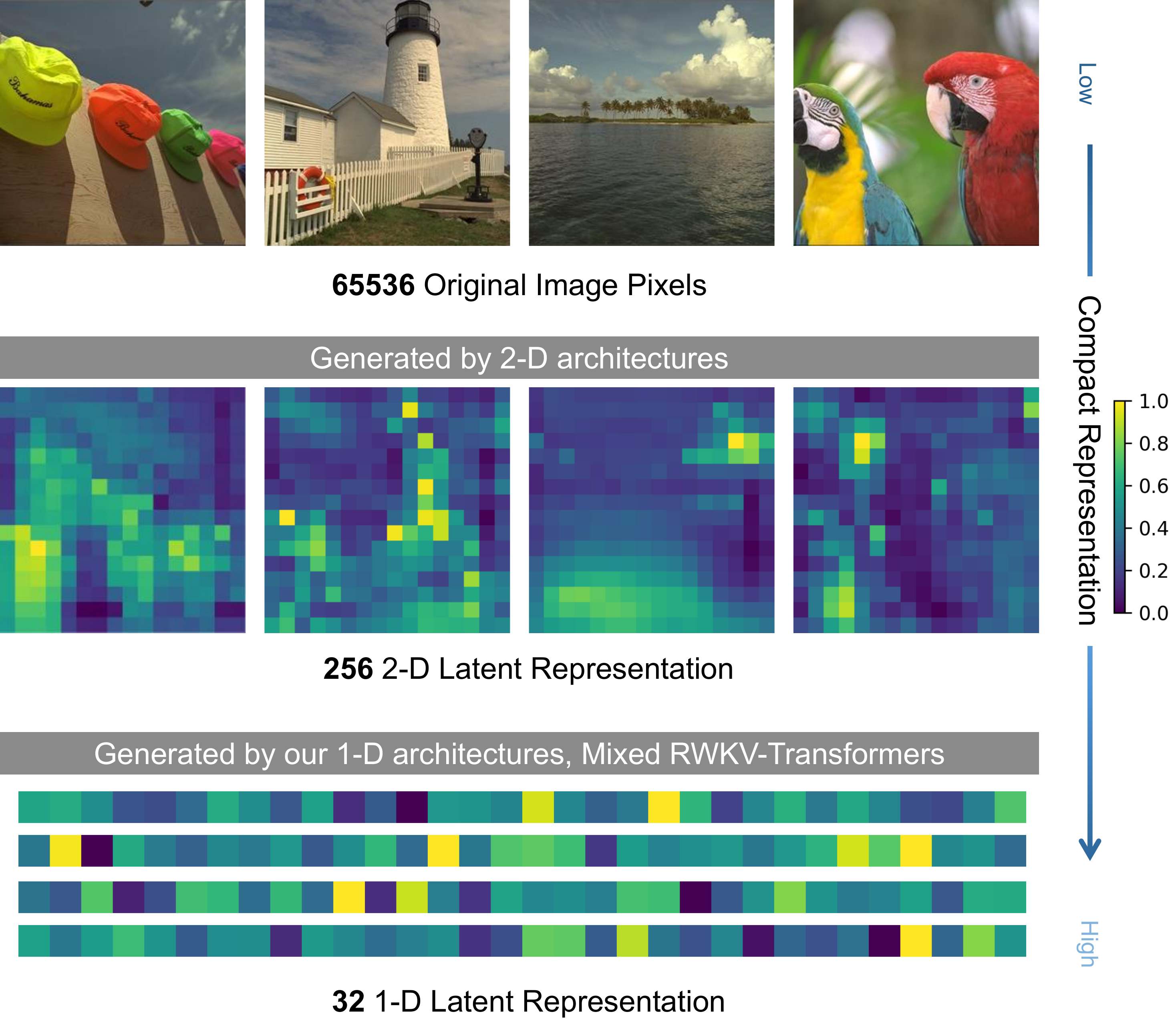} 
\caption{Visualization of original images and heatmaps of latent representations. 
2-D architectures\cite{illm} compress 65536 pixels into 256 latent features, 
while redundancy exists between these features. 
Our 1-D architectures generate more compact latent representations by 
producing 32 1-D latent features from 65536 pixels.}
\label{introduction}
\end{figure}

\begin{figure*}[t]
    \centering
    \includegraphics[width=0.82\textwidth]{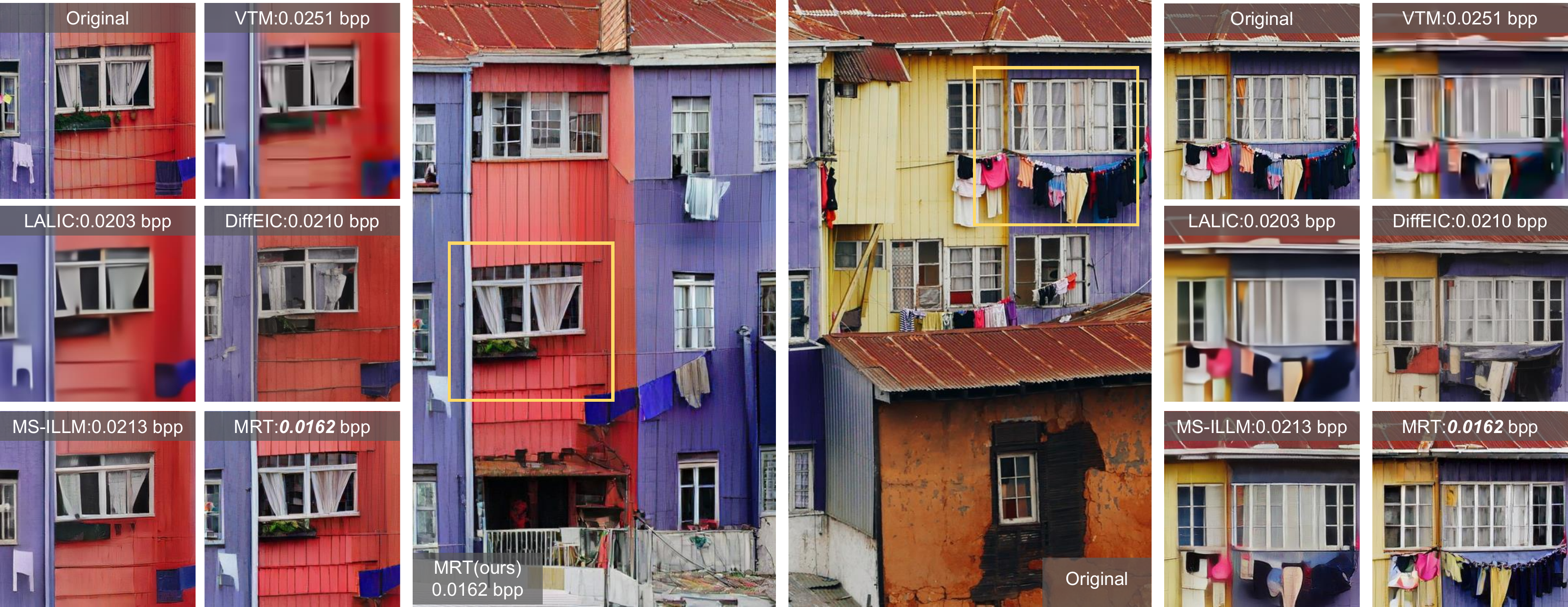} 
    \caption{Qualitative comparison of reconstruction quality across different 
    compression architectures. Our proposed MRT demonstrates superior visual fidelity 
    compared to VTM~\cite{VVC} and existing 2-D architectures including LALIC~\cite{lalic}, 
    DiffEIC~\cite{Diffeic} and MS-ILLM~\cite{illm}. MRT achieves better preservation 
    of fine details even at lower bitrates, while conventional 2-D architectures 
    exhibit significant degradation despite operating at higher bitrates.}
    \label{Introduction}
\end{figure*}

Most existing image compression methods rely on 2-D neural network architectures 
(e.g., CNNs and Swin Transformers) that compress images into 2-D latent representations 
(e.g., mapping a $256\times256$ image to a $16\times16$ latent feature map) to reduce 
redundancy in the representations, which are significantly 
more compact than the original image space and achieve superior rate-distortion performance 
compared to pixel-space codecs~\cite{VVC, lalic}.
However, these 2-D representations still exhibit redundancy among neighboring features, 
leading to suboptimal rate-distortion performance at extremely low bitrates. 
As illustrated in Figure~\ref{introduction}, the latent features produced by 2-D architectures 
tend to be highly correlated, limiting their efficiency in achieving truly compact representations~\cite{Titok}.

In response to such limitation, recent studies have investigated 1-D tokenizers based on Vision Transformers (ViTs)~\cite{Titok,FlexTok}, which transform images into a compact set of 1-D tokens for image generation (e.g., $32$ tokens for a $256\times256$ image as shown in Figure~\ref{introduction}). These 1-D architectures have demonstrated remarkable semantic compression capability, enabling compact yet expressive latent representations. However, ViT-based tokenizers are typically tailored for fixed-resolution image synthesis and require extensive fine-tuning to generalize across varying image sizes. In contrast, linear attention models such as RWKV~\cite{VRWKV,Restore-rwkv,lalic} exhibit robust extrapolation to variable sequence lengths and excel at modeling long-range dependencies. These observations motivate a fundamental question: \textit{How can we effectively combine the complementary strengths of ViTs and RWKV to achieve highly compact and  flexible representations for extreme image compression?}

To tackle these challenges, we propose a novel Mixed RWKV-Transformer (MRT) architecture for extremely compact latent representation in image compression. 
In MRT, each image is partitioned into fixed-size windows. 
A linear-attention-based RWKV block is employed to capture global dependencies across windows,
while ViT blocks are used to model local redundancies within each 
window. In contrast to previous codecs~\cite{illm,hific,GLC} which typically 
compress each window into 2-D feature maps, MRT encodes each window into a set 
of highly compact 1-D tokens, which are then aggregated to form a global latent 
representation of the entire image. Considering that existing 2-D compression 
models are not directly applicable to 1-D tokens due to their distinct 
structural characteristics, we design a dedicated RWKV Compression Model 
(RCM), which incorporates multiple Bi-RWKV blocks with spatial and channel mixing modules to effectively eliminate redundancy across both dimensions, enabling efficient compression of the 1D representations and achieving superior rate-distortion performance.

In summary, the main contributions of this work are as follows:
\begin{itemize}
\item We propose a novel MRT architecture for extreme image compression, which combines the global modeling capability of RWKV with the local representation strength of ViTs, representing an image as highly compact 1-D latent features.
\item We design a dedicated 1-D compression module, RCM, tailored for the intermediate 1-D features in MRT. By incorporating multiple Bi-RWKV blocks, RCM effectively estimates the entropy of latent representations and eliminates redundancy, further improving compression efficiency.
\item Extensive experimental evaluations demonstrate that MRT achieves significant bitrate savings of 
$43.75\%$, $30.59\%$ on the Kodak and CLIC2020 test datasets, respectively, while maintaining the same DISTS scores to state-of-the-art methods at extremely low bitrates. 
\end{itemize}


\section{Related Work}

\begin{figure*}[t]
    \centering
    \includegraphics[width=0.82\textwidth]{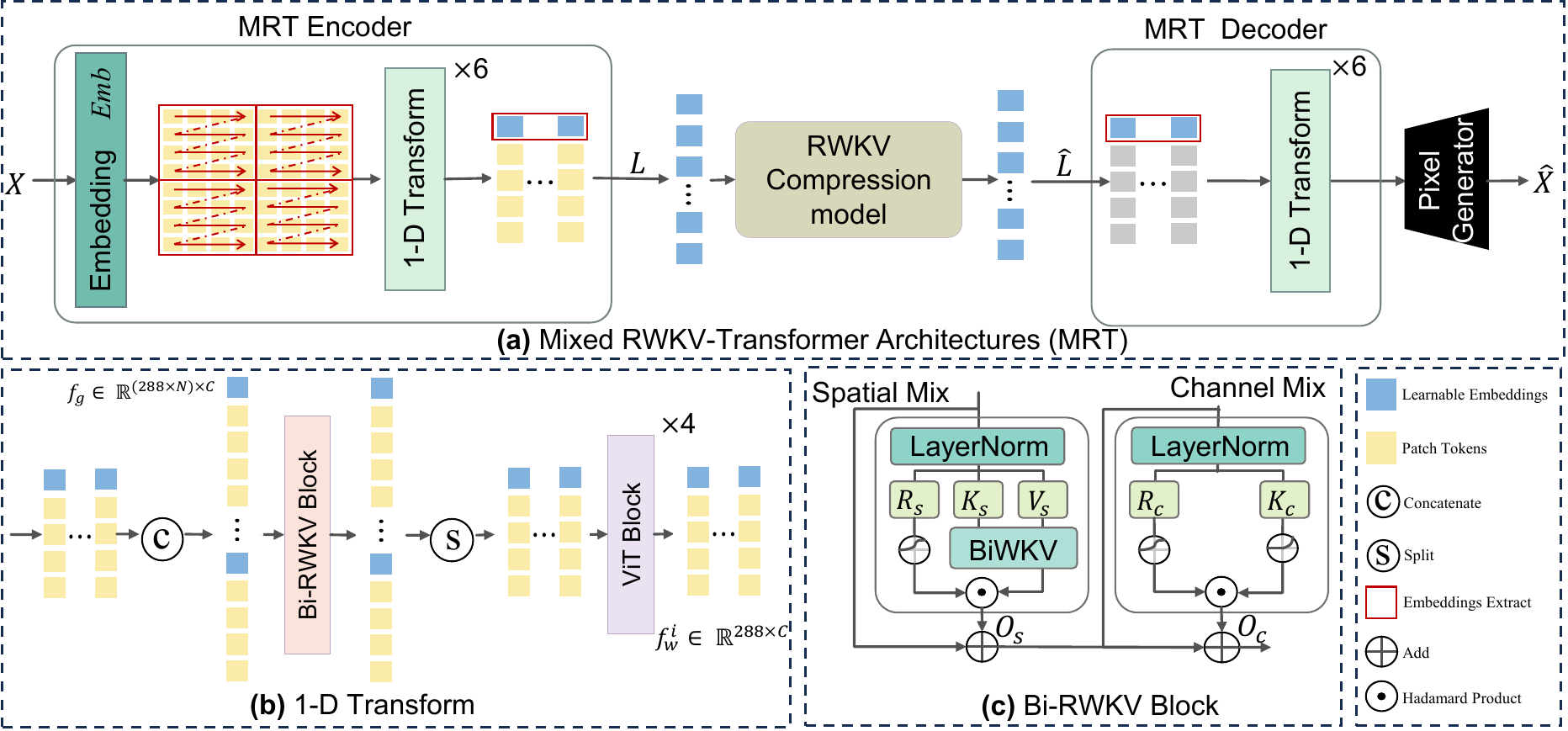} 
    \caption{(a) Overview of the proposed MRT, 
    which consists of a MRT encoder, RCM, a MRT decoder and 
    a pixel generator. 
    MRT encoder applies a stack of 1-D transformation layers to extract compact 1-D latent 
    representations, which are then compressed. The decoder mirrors this process to 
    reconstruct the image. (b) The 1-D transform alternates between Bi-RWKV and windowed 
    ViT blocks to capture both global and local dependencies. 
    (c) The Bi-RWKV block comprises spatial and channel mix modules, 
    each utilizing layer normalization, linear projections, and BiWKV attention for 
    efficient sequence modeling.}
    \label{Method}
\end{figure*}

\subsection{Generative Image Compression}
With the rapid development of deep learning~\cite{10003241,10608071}, generative models have achieved remarkable success in various computer vision tasks. Traditional hybrid coding frameworks ~\cite{VVC} and their deep-learning-enhanced variants~\cite{li2025ustc,10856241,li2024object,li2024loop} have significantly improved signal fidelity, yet they still suffer from severe quality degradation at extremely low bitrates, leading to unsatisfactory perceptual experience. Recent advances in generative models (e.g., tokenizers, GANs, diffusion models) have enabled 
superior perceptual quality in extreme image compression scenarios.
HIFIC~\cite{hific} first integrates conditional GANs into VAE frameworks, 
while MS-ILLM~\cite{illm} employs non-binary discriminators for improved statistical fidelity.
VQGAN~\cite{vqgan} demonstrates strong alignment between latent representations and human perceptual characteristics.
Recent works have integrated VQVAE tokenizers into compression frameworks, with VQ-means~\cite{finetuneVQGAN} 
introducing K-means clustering and UIGC~\cite{uigc} achieving better rate-perception performance through 
transformer-based token regeneration.
GLC~\cite{GLC} further reduces redundancy through latent space compression.
Diffusion-based approaches~\cite{RDEIC, yang2024lossy, körber2024percosdopenperceptual, jiang2023multi} have also shown promise in generative compression.
However, existing methods rely on 2-D architectures that limit compression efficiency due to spatial redundancy.
To address this limitation, we propose a 1-D architecture that encodes images into compact 1-D latent representations, enabling more effective extreme image compression.

\subsection{Linear Attention}
Among emerging linear attention models, Receptance Weighted Key Value (RWKV)~\cite{VRWKV, yuan2024mamba} and Mamba~\cite{mambda, zhu2024vision} have demonstrated significant potential for capturing long-range dependencies. This work focuses on RWKV as the foundational component of our MRT architecture.
Originally developed for natural language processing~\cite{rwkv}, RWKV employs a novel WKV attention mechanism 
that excels at modeling long-range dependencies across extended sequences while maintaining linear computational complexity.
Recent work has shown RWKV's effectiveness in computer vision tasks, particularly in scenarios requiring global context understanding. Vision RWKV (VRWKV)~\cite{VRWKV} achieves 
comparable performance to Vision Transformers in image synthesis by effectively capturing long-range spatial relationships.
Restore-RWKV~\cite{Restore-rwkv} has established state-of-the-art benchmarks in image restoration by leveraging RWKV's ability to model dependencies across distant image regions, 
and LALIC~\cite{lalic} has successfully integrated RWKV into image compression frameworks, 
demonstrating superior compression performance through enhanced long-range dependency modeling. 

\section{Method}
\subsection{Mixed RWKV-Transformers Architecture}

In this paper, we propose a novel framework, MRT, which transforms images into compact 1-D latent representations for more efficient image compression. MRT is composed of three key modules: MRT encoder, RCM, and MRT decoder. The encoder employs multiple 1-D transformation layers designed to capture both inter-window and intra-window dependencies, followed by RCM for latent space compression. The decoder adopts a symmetric architecture as the encoder with a pixel generator to reconstruct images from latent representations. The subsequent sections provide a detailed exposition of each module.

\subsubsection{Encoding}
Given an input image $X\in\mathbb{R}^{3\times H\times W}$, we first use a CNN layer 
to embed it into patch tokens $Emb(X)\in\mathbb{R}^{w\times h\times c}$, where $w=H/16$ and $h=W/16$. The 
embedded patch tokens are then partitioned into $N$ non-overlapping $16\times16$ windows, 
each of which is flattened to a token sequence $L_i\in\mathbb{R}^{256\times c}$. To compress images into highly compact 1-D latent representations, we follow~\cite{Titok} and append a learnable latent embedding $L_{latent}\in\mathbb{R}^{32\times c}$ to each window, forming an extended token sequence $L_w^i\in\mathbb{R}^{288\times c}$, as illustrated in Figure~\ref{Method}(a), where blue squares represent the learnable latent embeddings.

After that, MRT adopts a 1-D transform module to jointly processe all window tokens, as illustrated in Figure~\ref{Method} (b). The process consists of three steps: (1) All windows tokens are concatenated together to form a global representation $f_g\in\mathbb{R}^{(288\times N)\times c}$, enabling global context modeling; (2) A Bi-RWKV is applied to capture long-range dependencies across windows within $f_g$; (3) The transformed global representation is reshaped back to individual windows and passed through ViT blocks for local spatial modeling, yielding $f_w^i\in\mathbb{R}^{288\times c}$.

Finally, we extract the learnable embeddings $L_{latent}$ from each processed window and concatenate them to form a highly compact 1-D representation $L\in\mathbb{R}^{(N\times 32)\times c}$, which serves as the final latent representation of the entire image and is subsequently compressed by the RCM, as illustrated in Figure~\ref{Method} (a).

\subsubsection{Decoding}
On the decoder side, given the quantized representation $\hat{L}\in\mathbb{R}^{(N\times 32)\times c}$, 
we partition it into $N$ segments. Each segment is concatenated with mask features, 
which share the same shape as flattened features $L^i_w\in\mathbb{R}^{288\times c}$, to reconstruct window 
representation $\hat{L}^i_w\in\mathbb{R}^{288\times c}$. The windows are processed through 
1-D transformation layers: Bi-RWKV for global dependencies, then ViT for local 
relationships. After removing embeddings, features are rearranged to 2-D map 
$f\in\mathbb{R}^{w\times h\times c}$, and pixel generator $G$ produces output 
$\hat{X}\in\mathbb{R}^{3\times H\times W}$.

\begin{figure}[t]
    \centering
    \includegraphics[width=0.9\columnwidth]{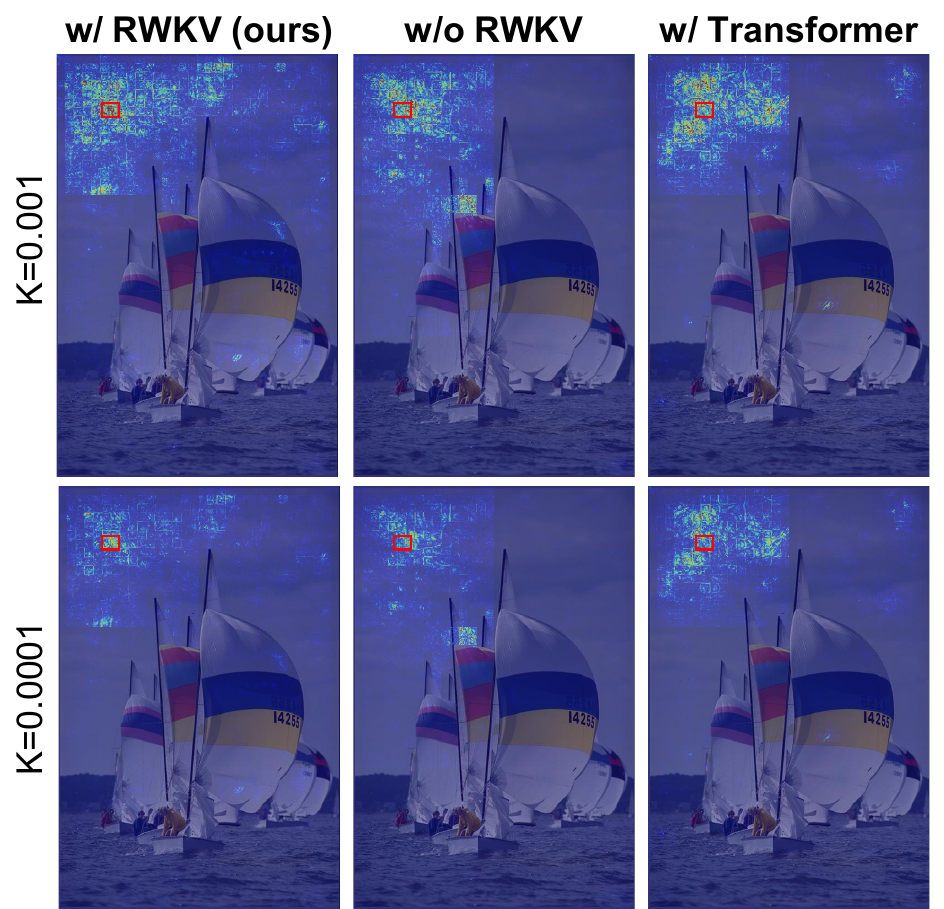} 
    \caption{Effective receptive field visualization for different architectural 
    configurations on \textit{kodim09}. 
    Columns: our proposed model with global Bi-RWKV blocks, 
    model without cross-window dependency modeling, and 
    model with global ViT blocks. Rows: gradient thresholds (0.0001 and 0.001).}
    \label{Visualization}
\end{figure}

\begin{figure*}[t]
    \centering
    \includegraphics[width=0.84\textwidth]{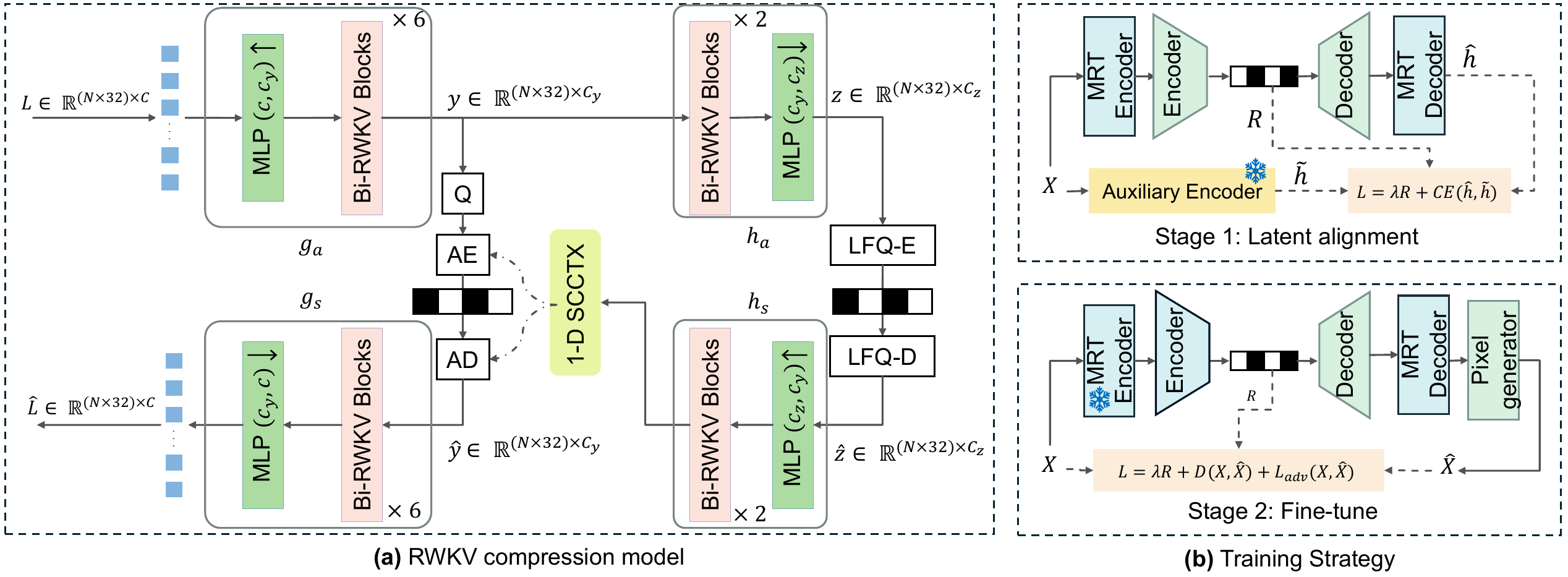} 
    \caption{
        (a) RWKV compression model architecture. 
        MLP$(c_1, c_2)\uparrow$ and MLP$(c_2, c_1)\downarrow$ denote 
        dimensionality expansion and reduction, respectively.
        AE, AD, and Q represent arithmetic encoding, arithmetic decoding, and quantization.
        LFQ-E, LFQ-D stand for look-up free (LFQ) encoding and decoding.
        1-D SCCTX denotes the 1-D Spatial-Channel Context Module.
        (b) Two-stage training strategy. 
        Stage 1: auxiliary encoder aligns quantized codes with discrete targets via cross-entropy loss. 
        Stage 2: RCM and MRT decoder are optimized end-to-end using pixel-level and perceptual losses.
    }
    \label{Compression}
\end{figure*}

\subsubsection{Bi-RWKV Block}
To efficiently capture cross-window dependencies and support variable-resolution images, we adopt a Bi-RWKV block that
consists of two primary components: spatial mix and channel mix. Firstly, the spatial mix module applies layer normalization and three linear projections to generate receptance $R_s$, key $K_s$, and value $V_s$. After that, given $K_s$ and $V_s$, global attention is computed using the Bi-directional Weighted Key-Value (BiWKV) mechanism, modulated by sigmoid-gated receptance $\theta(R_s)$ to produce spatial output $O_s$. The BiWKV attention for the $t$-th token is defined as:
\begin{equation}
\mathrm{wkv}_t =
\frac{
\sum\limits_{i=1,\,i\neq t}^{T} e^{-\frac{|t-i|-1}{T} \cdot w + k_i} v_i + e^{u + k_t} v_t
}{
\sum\limits_{i=1,\,i\neq t}^{T} e^{-\frac{|t-i|-1}{T} \cdot w + k_i} + e^{u + k_t}
}
\end{equation}
where $k_i$ and $v_i$ denote the key and value of the $i$-th token, and $T$ represents the sequence length of 1-D representations. This enables linear computational complexity for long-range dependencies. Finally, the channel mix module takes $O_s$ as input and adopts two linear projections to produce receptance $R_c$ and key $K_c$. A squared ReLU activation is applied to $K_c$, and the final channel output $O_c$ is obtained via sigmoid-gated modulation, analogous to the spatial pathway.

To evaluate Bi-RWKV's effectiveness in modeling long-range dependencies, we conduct an effective receptive field (ERF) analysis following~\cite{tcm}.
We compare Bi-RWKV against two baselines: (1) replacing global Bi-RWKV blocks with global RoPE-ViT blocks and (2) removing global Bi-RWKV blocks entirely.
The ERF is computed as the absolute gradient magnitude with respect to a target region (marked by a red square).
We clip the top $K$ gradients to better visualize global gradients.
As shown in Figure~\ref{Visualization}, Bi-RWKV achieves superior global dependency capture with broader and more uniform gradient distribution across the entire image. In contrast, models without Bi-RWKV blocks show localized attention patterns concentrated around the target area, while models with global RoPE-ViT blocks exhibit incomplete cross-window interaction.
The visualization confirms that Bi-RWKV effectively models long-range dependencies for comprehensive spatial coverage in image compression.

\subsection{RWKV Compression Model}
Compared to conventional 2-D latent representations produced by 2-D 
architectures~\cite{illm,hific,GLC}, our proposed 1-D latent representations are highly 
compact. However, due to the absence of spatial priors typically leveraged in 2-D representations, 1-D latent do not support straightforward downsampling for hyperprior extraction. 
To address this challenge, we propose a dedicated RWKV compression model (RCM) that directly 
operates on the 1-D latent features.
As depicted in Figure~\ref{Compression}, the 1-D latent representations are first projected 
into a higher-dimensional space and then transformed into a latent 
variable $y \in \mathbb{R}^{(N \times 32) \times c_y}$, where $c_y = 320$. 
The variable $y$ is subsequently quantized to obtain $\hat{y}$. 
To model the distributional properties of $y$, a hyper network is employed to
 generate hyper representations $z \in \mathbb{R}^{(N \times 32) \times c_z}$ by 
 processing $y$ through two stacked Bi-RWKV blocks, followed by a multilayer 
 perceptron (MLP) that projects the output into a lower-dimensional hyper space.
To maximize codebook utilization, we employ Look-up Free Quantization (LFQ) to quantize the hyper representations, 
rather than the prior VQ-based hyper modules~\cite{GLC}. 
We define LFQ as:
\begin{equation}
\hat{z} = q(z) = 2 \cdot \mathbb{I}[z \geq 0] - 1.
\end{equation}
In our implementation, we set $c_z = 14$, resulting in a codebook of size $16384$ for 
the binary hyper representations. 
To further enhance the compression efficiency, we introduce a 1-D Spatial-Channel Context 
Module (SCCTX) to estimate the entropy of the quantized hyper representations. SCCTX facilitates more accurate probability modeling for entropy coding, thereby reducing the overall bitrate. Finally, the quantized latent 
representations $\hat{y}$ are passed to the MRT decoder to reconstruct the 
compact 1-D latent features $\hat{L} \in \mathbb{R}^{(N \times 32) \times c}$.

\subsection{Two-Stage Training Strategy}
Inspired by recent works~\cite{Titok, FlowMo}, we train MRT in a two-stage manner, including latent alignment and decoder fine-tuning, as shown in Figure~\ref{Compression}(b).

\subsubsection{Latent Alignment.}
Instead of training the entire model with pixel-level
loss functions, we align the codes $\hat{h}$ with the discrete
codes $\tilde{h}$ generated by a pre-trained VQGAN model~\cite{maskgit} using cross-entropy loss denoted 
$CE(\hat{h},\tilde{h})$ to measure the reconstruction loss. 
In order to make the RCM module more effective in compressing the intermediate 
1-D latent, multiple dedicated loss function are also introduced. 
Firstly, for stability, we also introduce a space alignment loss to minimize the 
distortion from the RCM~\cite{Diffeic}:
\begin{equation}
    \mathcal{L}_{\mathrm{latent}} = \mathcal{L}_{MSE}(\hat{L}, L)
\end{equation}
Additionally, following~\cite{LFQ}, an auxiliary loss is introduced to maximize the 
usage of the LFQ codebook, which is defined as:
\begin{equation}
\mathcal{L}_{\mathrm{ent}} = \mathbb{E}\left[ H(q(\hat{z})) - H\left(\mathbb{E}[q(\hat{z})]\right) \right], \\
\end{equation}
\begin{equation}
\mathcal{L}_{\mathrm{commit}} = \mathbb{E}\left[ \left\| \hat{z} - q(\hat{z}) \right\|_2^2 \right].
\end{equation}
\begin{equation}
    \mathcal{L}_{\mathrm{LFQ}} = \lambda_{ent}\cdot\mathcal{L}_{\mathrm{ent}} + \lambda_{commit}\cdot\mathcal{L}_{\mathrm{commit}}.
\end{equation}
where $\lambda_{ent}$ and $\lambda_{commit}$ are the weights of the entropy loss and commitment loss, respectively.
We set $\lambda_{ent}=0.25$ and $\lambda_{commit}=0.00625$.
Finally, we optimize the estimated entropy of
the representation $y$, resulting in $R$. Overall, we train MRT encoder and MRT decoder and RCM with the following loss function:
\begin{equation}
\mathcal{L}_{\mathrm{stage1}} = \mathcal{L}_{\mathrm{CE}} + \mathcal{L}_{\mathrm{LFQ}} + \mathcal{R}(\hat{y}) + \mathcal{L}_{\mathrm{latent}}
\end{equation}
In stage 1, we do not modify the lambda of rates to achieve different bitrates.

\begin{figure*}[t]
    \centering
    \includegraphics[width=0.82\textwidth]{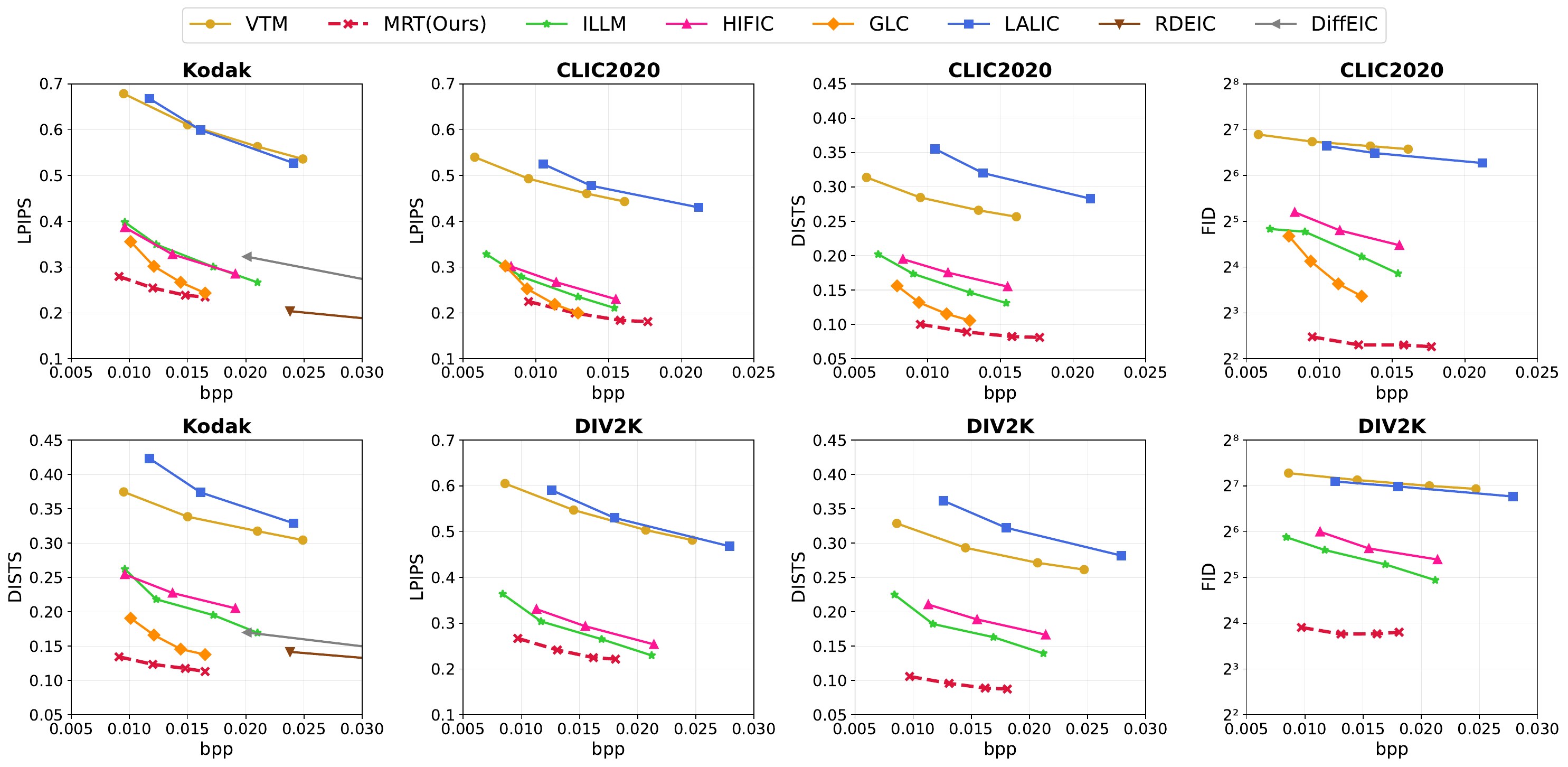} 
    \caption{Rate-distortion curves on the Kodak, the CLIC2020 and the DIV2K datasets.}
    \label{RD-curve}
\end{figure*}
\begin{figure*}[t]
    \centering
    \includegraphics[width=0.85\textwidth]{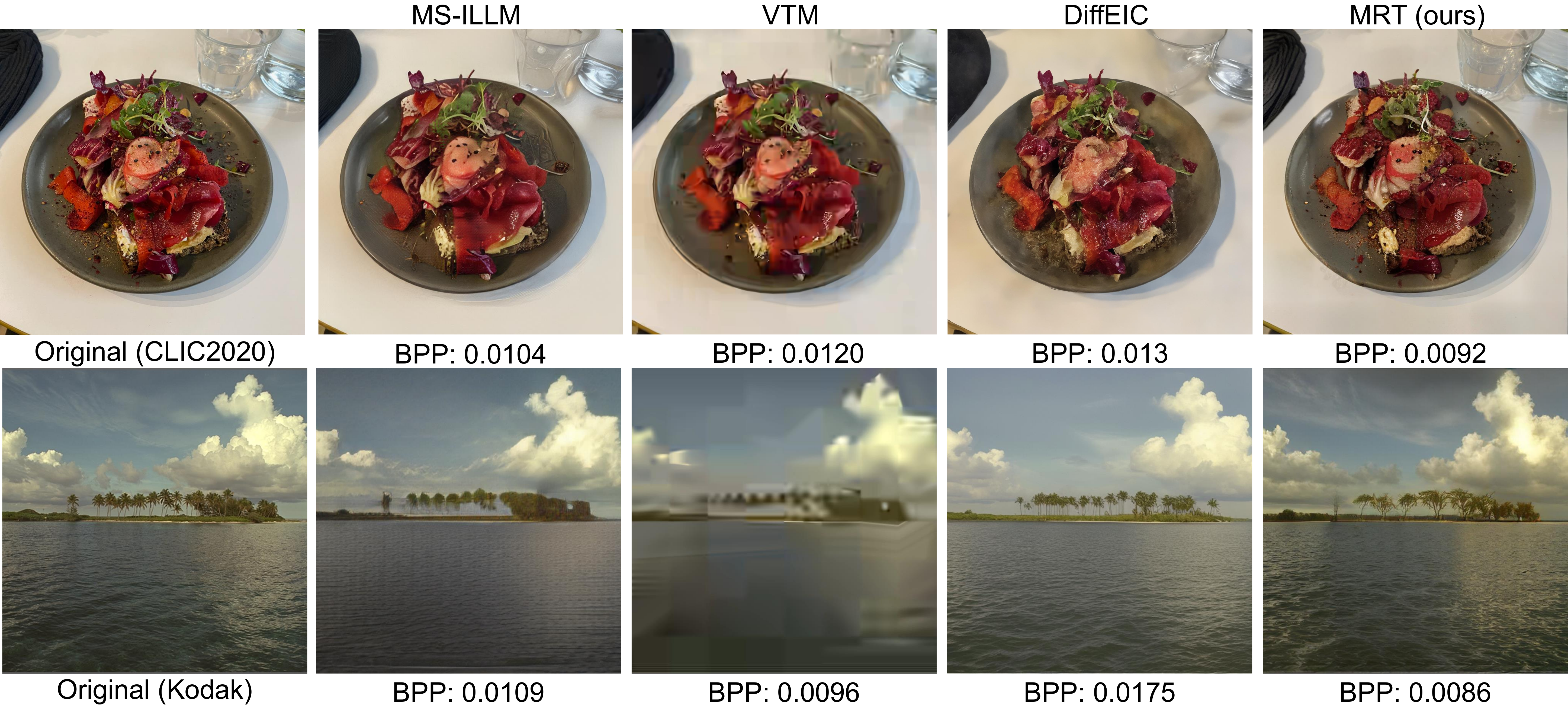} 
    \caption{Qualitative examples of different methods on Kodak and CLIC2020. Zoom in for better visualization.}
    \label{Qualitative}
\end{figure*}

\subsubsection{Decoder Fine-tuning.}
In the decoder fine-tuning stage, we jointly optimize the RCM, 
MRT decoder and pixel generator using pixel-level supervision. Specifically, we employ a combination of 
pixel-wise $\ell_1$ loss, perceptual loss~\cite{Vgg}, and adversarial loss~\cite{Titok} to 
enhance both the fidelity and perceptual quality of the reconstructed images. 
The overall objective for this stage is formulated as:
\begin{equation}
\mathcal{L}_{\mathrm{stage2}} = \mathcal{L}_{1} + \mathcal{L}_{\mathrm{perceptual}} + \lambda_{\mathrm{adv}} \mathcal{L}_{\mathrm{adv}} + \lambda \mathcal{R}(\hat{y})
\end{equation}
where $\mathcal{L}_{1}$ denotes the pixel-wise $\ell_1$ loss, $\mathcal{L}_{\mathrm{perceptual}}$ is 
the widely used VGG-based perceptual loss~\cite{Vgg}, 
$\mathcal{L}_{\mathrm{adv}}$ is the adversarial loss, and $\mathcal{R}$ represents exstimated bitrates. The hyperparameters $\lambda_{\mathrm{adv}}$ and $\lambda$ are used to balance the contributions of the adversarial loss and the rate, respectively. In our experiments, we set $\lambda_{\mathrm{adv}} = 0.05$ by default.

\section{Experiments}
\subsection{Implementation Details}

\subsubsection{Training Details.}
Our MRT model is trained on the OpenImages training set~\cite{openimages} following a two-stage training strategy. 
In the first stage, we train the model using randomly cropped $512\times512$ image patches with a batch size of 4, 
initialized with a pre-trained 1-D tokenizer~\cite{Titok} to accelerate convergence. 
In the second stage, we fine-tune the MRT decoder, RCM, and pixel generator using the same patch size and batch size. 
To achieve different rate-distortion trade-offs, we set the rate-distortion parameter $\lambda$ to $\{20, 10, 5, 2.5\}$. 
All models are trained on a single NVIDIA RTX 5090 GPU.

\subsubsection{Evaluation Datasets.}
We evaluate our MRT model on three widely-used image compression benchmarks: the Kodak
dataset~\cite{kodak}, CLIC2020 test set~\cite{CLIC2020}, and DIV2K validation 
set~\cite{DIV2K}. We utilize full-resolution images to assess performance across diverse image characteristics. 
More results are provided in the supplementary material.

\subsubsection{Evaluation Metrics.}
We employ multiple evaluation metrics to assess performance. 
For reconstruction fidelity, we use perceptual metrics LPIPS~\cite{LPIPS} and DISTS~\cite{DISTS}. 
Generation realism is quantified using FID~\cite{FID}. 
Bitrate is calculated in bits per pixel (bpp). Following~\cite{hific}, we do not report FID for Kodak due to its limited size. 
Additional PSNR and MS-SSIM~\cite{MS-SSIM} results are provided in supplementary material.

\subsubsection{Comparison Methods.}
We compare our method with state-of-the-art image compression approaches. 
Traditional baselines include VTM-23.11~\cite{VVC} and the RWKV-based neural codec LALIC~\cite{lalic}. 
For generative compression, we evaluate against HiFiC~\cite{hific}, MS-ILLM~\cite{illm}, 
GLC~\cite{GLC}, and Diffusion-based codecs DiffEIC~\cite{Diffeic}, RDEIC~\cite{RDEIC}.

\subsection{Main Results}
\subsubsection{Quantitative Evaluation}
Figure~\ref{RD-curve} shows rate-distortion curves across datasets and metrics~\cite{bjontegaard2001calculation}. 
MRT achieves the best rate-distortion trade-offs on three datasets (Kodak, CLIC2020, DIV2K) and three metrics (LPIPS, DISTS, FID). 
On Kodak, MRT achieves $19.82\%$ and $43.75\%$ bitrate savings for LPIPS and DISTS compared to GLC, outperforming diffusion-based methods. 
On CLIC2020, MRT maintains $30.59\%$ bitrate saving for DISTS. MRT achieves competitive FID performance on both DIV2K and CLIC2020 datasets.

\subsubsection{Qualitative Evaluation}
Figure~\ref{Qualitative} shows reconstruction quality comparisons. 
On CLIC2020 image, MRT (0.0092 bpp) maintains sharp textures and fine details, while MS-ILLM shows texture softening, VTM exhibits blurring, and DiffEIC loses details. 
On Kodak image, MRT (0.0086 bpp) produces clear reconstructions and preserves textures. 
VTM suffers from banding artifacts and blurring, while MS-ILLM shows reduced sharpness. 
DiffEIC requires higher bitrate with mismatch. MRT generates more visually pleasing results at extreme bitrates.

\subsection{Ablation Study}
In this section, we conduct ablation studies to analyze the contribution of MRT,
Bi-RWKV compression model and look-up free quantization. All model variations are 
evaluated on the Kodak, CLIC2020 and DIV2K datasets using DISTS.
\subsubsection{Mixed RWKV-Transformer Architecture.}
We evaluate the contribution of global Bi-RWKV blocks by removing them and 
keeping only local ViT blocks (denoted as $\textit{MRT w/o RWKV}$). 
Table~\ref{ablation} shows bitrate increases of $9.62\%$, $32.41\%$, 
and $25.76\%$ on Kodak, CLIC2020, and DIV2K datasets, respectively. 
The more pronounced performance degradation on CLIC2020 and DIV2K (high-resolution images) 
highlights MRT's superior capability in compressing high-resolution images without 
requiring fine-tuning.

\begin{table}
    
    \centering
    \begin{tblr}{
      column{even} = {c},
      column{3} = {c},
      vline{2} = {-}{},
      hline{1-2,6} = {-}{},
    }
    Model variants  & Kodak & CLIC2020 & DIV2K \\
    MRT w/o RWKV & 9.62\% & 32.41\% & 25.76\% \\
    RCM w/ ViT & 6.99\% & 8.61\% & 9.08\% \\
    RCM w/ VQ & 34.52\% & 40.66\% & 40.40\% \\
    $\textbf{Ours}$ & 0.0\% & 0.0\% & 0.0\% \\
    \end{tblr}
    \caption{ BD-Rate on DISTS metric of different model variants. We use the proposed methods as the anchor, denoted as $\textbf{Ours}$.}
    \label{ablation}

\end{table}

\begin{table}
    \centering
    \begin{tabular}{l|ccc} 
    \hline
    Quantization method & Kodak & CLIC2020 & DIV2K  \\ 
    \hline
    MRT w/ VQ & 5.42 & 5.65 & 5.68 \\
    MRT w/ LFQ & 7.59 & 10.38 & 10.39 \\
    \hline
    \end{tabular}
    \caption{Entropy of the codebook of different quantization methods. Larger entropy indicates more effective usage of the codebook.}
    \label{codebook}
\end{table}

\subsubsection{RWKV Compression Model.}
We replace Bi-RWKV blocks with RoPE-ViT blocks in RCM (denoted as $\textit{RCM w/ ViT}$). Table~\ref{ablation} shows bitrate increases of $6.99\%$, $8.61\%$, and $9.08\%$ on the three datasets, indicating that Bi-RWKV blocks are more effective for modeling long-range dependencies in the compression domain.

\subsubsection{Look-up Free Quantization.}
We compare LFQ against conventional vector quantization (denoted as $\textit{RCM w/ VQ}$).
As observed in Table~\ref{ablation}, replacing VQ with LFQ results in significant bitrate reductions of $34.52\%$, $40.68\%$, and $42.52\%$ across three benchmarks. 
Additionally, as illustrated in Table~\ref{codebook}, LFQ achieves higher entropy than VQ, indicating superior codebook utilization. This advantage facilitates more complex distribution modeling and enable more efficient compression in the 1-D latent space.

\begin{table}
    \centering
    \begin{tblr}{
      row{2} = {c},
      column{4} = {c},
      cell{1}{1} = {r=2}{},
      cell{1}{2} = {c=2}{c},
      cell{1}{4} = {r=2}{},
      cell{3}{2} = {c},
      cell{3}{3} = {c},
      cell{4}{2} = {c},
      cell{4}{3} = {c},
      cell{5}{2} = {c},
      cell{5}{3} = {c},
      vline{2} = {1-5}{},
      hline{1,3,6} = {-}{},
    }
    Model variants & Latency (s) &   & BD-DISTS \\
      & Enc. & Dec. &   \\
    MS-ILLM & $\textbf{0.0350}$ & $\textbf{0.0304}$ & $0.00\%$ \\
    DiffEIC & $0.1456$ & $3.9079$ & $-34.52\%$ \\
    \textbf{Ours} & $0.1641$ & $0.1909$ & $\textbf{-77.64\%}$ 
    \end{tblr}
    \caption{Complexity analysis and BD-DISTS of different methods. All tests are conducted on the Kodak dataset with a single NVIDIA A100 GPU.}
    \label{complexity}
\end{table}

\section{Complexity Analysis}
We analyze the computational complexity of MRT. As shown in Table~\ref{complexity}, MRT has longer encoding and decoding time than traditional 2-D methods (e.g., MS-ILLM), but offers superior performance that justifies this trade-off. Compared to diffusion-based methods (e.g., DiffEIC), MRT achieves better rate-distortion performance while maintaining similar encoding time and significantly reducing decoding time. Additional results are provided in the supplementary material.
 
\section{Conclusion}
We propose a novel Mixed RWKV-Transformer (MRT) architecture for extreme image compression. 
By combining ViT blocks for local modeling and Bi-RWKV blocks for global modeling, MRT efficiently encodes images into compact 1-D representations. 
Additionally, we design a dedicated RWKV compression model (RCM) to compress these representations effectively. 
Experimental results demonstrate that MRT achieves superior rate-distortion performance across multiple datasets at extreme bitrates below 0.02 bpp. 
Future work will focus on optimizing the architectural design and training strategies to achieve comparable performance while reducing computational complexity.

\section*{Acknowledgment}
This work was supported in part by the National Key R\&D Program of China (2023YFA1008500), the National Natural Science Foundation of China (NSFC) under grants U22B2035 and 62502116, and China Post-Doctoral Science Foundation under Grant 2025M774315.

\bibliography{aaai2026}

\newpage

\appendix

\section{Appendix1: Model Architecture}

Our proposed model comprises two main components: a Mixed RWKV-Transformer 
(MRT), and a dedicated RWKV compression model (RCM).
In this section, we provide a detailed description of the proposed MRT and RCM.

\subsection{Mixed RWKV-Transformer}
\begin{figure*}[t]
    \centering
    \includegraphics[width=0.9\textwidth]{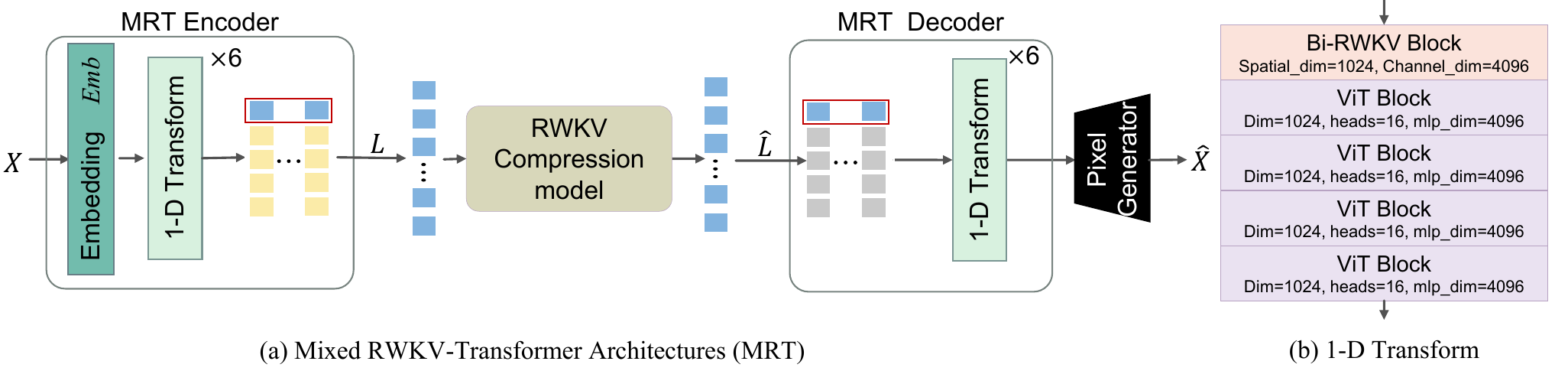} 
    \caption{Overview of Mixed RWKV-Transformer and hyperparameters of 1-D Transform.}
    \label{1D_Transform}
\end{figure*}

We present the detailed architecture specifications of the 1D transform with comprehensive hyperparameters.
As depicted in Figure~\ref{1D_Transform}, the 1D-transform architecture comprises 1 Bi-RWKV block
and 4 ViT blocks, all sharing identical dimensions with $d = 1024$ and the same ratio of $r = 4$.

\subsection{RWKV Compression Model}
\begin{figure*}[t]
    \centering
    \includegraphics[width=0.9\textwidth]{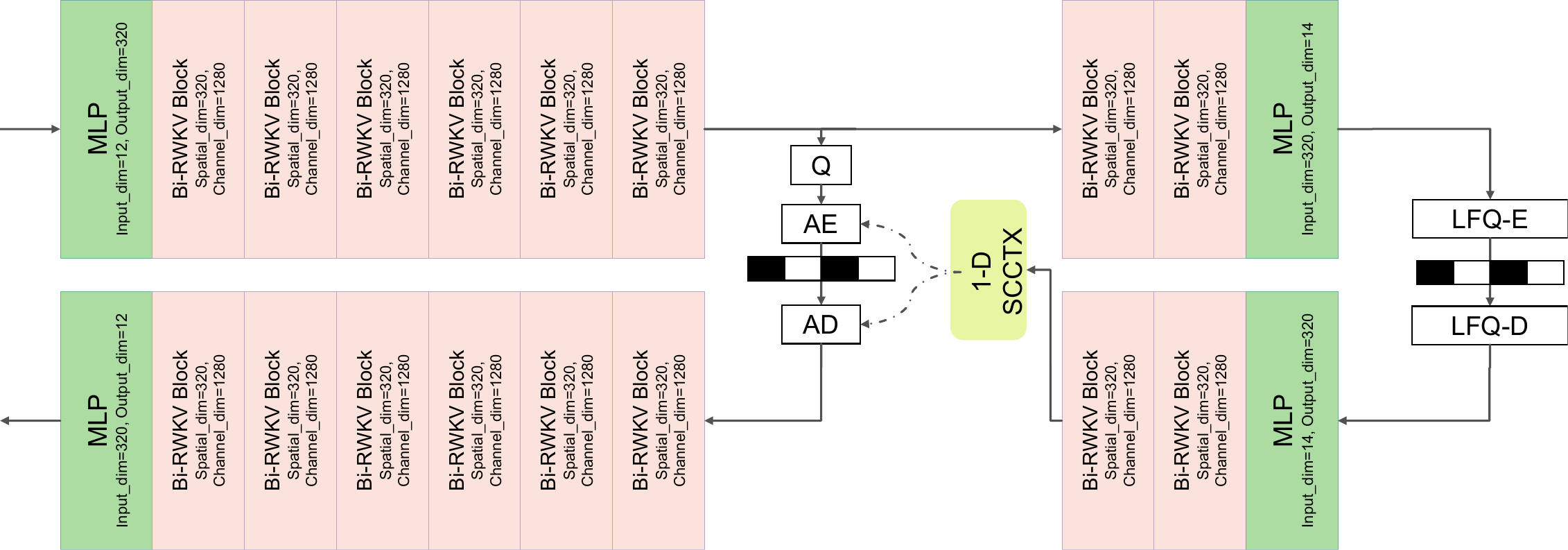} 
    \caption{Hyperparameters of the RWKV compression model.}
    \label{RCM_appendix}
\end{figure*}

As illustrated in Figure~\ref{RCM_appendix}, the RWKV compression model utilizes multiple Bi-RWKV blocks to effectively capture both spatial and channel dependencies, with all 
Bi-RWKV blocks sharing identical dimensions of $d = 320$ and a ratio of $r = 4$.
For the 1D SCCTX component, we adopt the entropy model from TCM~\cite{tcm} to model the distribution of the 1D latent features, 
simply replacing Conv blocks with Bi-RWKV blocks.

\section{Appendix2: Experiments}

\subsection{Training Details}
We train our model using the Open Image v4 training dataset~\cite{openimages} with a carefully designed two-stage 
training strategy. In the first stage, we employ the AdamW optimizer~\cite{adamw} with a weight decay of $1\times10^{-4}$ and a learning rate of $1\times10^{-4}$, 
incorporating 10,000 warmup steps followed by a cosine decay schedule. In the second stage, we maintain identical optimizer settings including learning rate, 
warmup steps, and decay schedule. To enhance training stability, we introduce the discriminator component only after 20,000 training steps.
For both training stages, we utilize a batch size of 4, an exponential moving average (EMA) decay of 0.999, and employ mixed-precision training.

\subsection{Evaluation Details}

\subsubsection{Evaluation Metrics}
We evaluate the performance of our model on the three widely used datasets: CLIC2020
test set~\cite{CLIC2020}, Kodak~\cite{kodak}, and DIV2K vaild set~\cite{DIV2K} using the following metrics:
LPIPS~\cite{LPIPS}, DISTS~\cite{DISTS}, and FSIM~\cite{FID}. Following HIFIC~\cite{hific},
we evaluate fid by splitting the images into 256 overlapped patches, noting that 
fid is not computed on the Kodak dataset as kodak only contains limited number of images.
All metrics are computed on the full resolution images of three datasets.

\subsubsection{Evaluation Methods}
We evaluate the performance of LALIC~\cite{lalic}, MS-ILLM~\cite{illm}, and HIFIC~\cite{hific} 
by utilizing the official released checkpoints and fine-tuning their pretrained models to extremely low
bitrates. As for RDEIC~\cite{RDEIC} and DiffEIC~\cite{Diffeic}, we employ the official released checkpoints 
for evaluation.
For GLC~\cite{GLC}, we obtain the evaluation results through personal communication with the authors.

\subsection{Redundancy Analysis}
We present a redundancy analysis comparing 2D representations from MS-ILLM~\cite{illm} and 1D representations from 
MRT in Table~\ref{redundancy}.
Spatial redundancy is measured via cosine similarity and $L_2$ distance, while channel redundancy is measured 
via mean feature correlation (MFC).
Results show MRT achieves lower spatial and channel redundancy than MS-ILLM, indicating more compact and information-efficient representations.

\subsection{Complexity Analysis}
We provide the computational complexity analysis of different methods in Table~\ref{complexity}.
MRT outperforms the 2D architecture method MS-ILLM~\cite{illm} on BD-DISTS metrics, despite having more
model parameters and higher computational complexity. Additionally, MRT achieves
better BD-DISTS performance than DiffEIC~\cite{Diffeic} while requiring fewer model parameters and maintaining competitive encoding time with significantly reduced decoding time.

\subsection{Quantitative Results}
\begin{figure*}[t]
    \centering
    \includegraphics[width=0.9\textwidth]{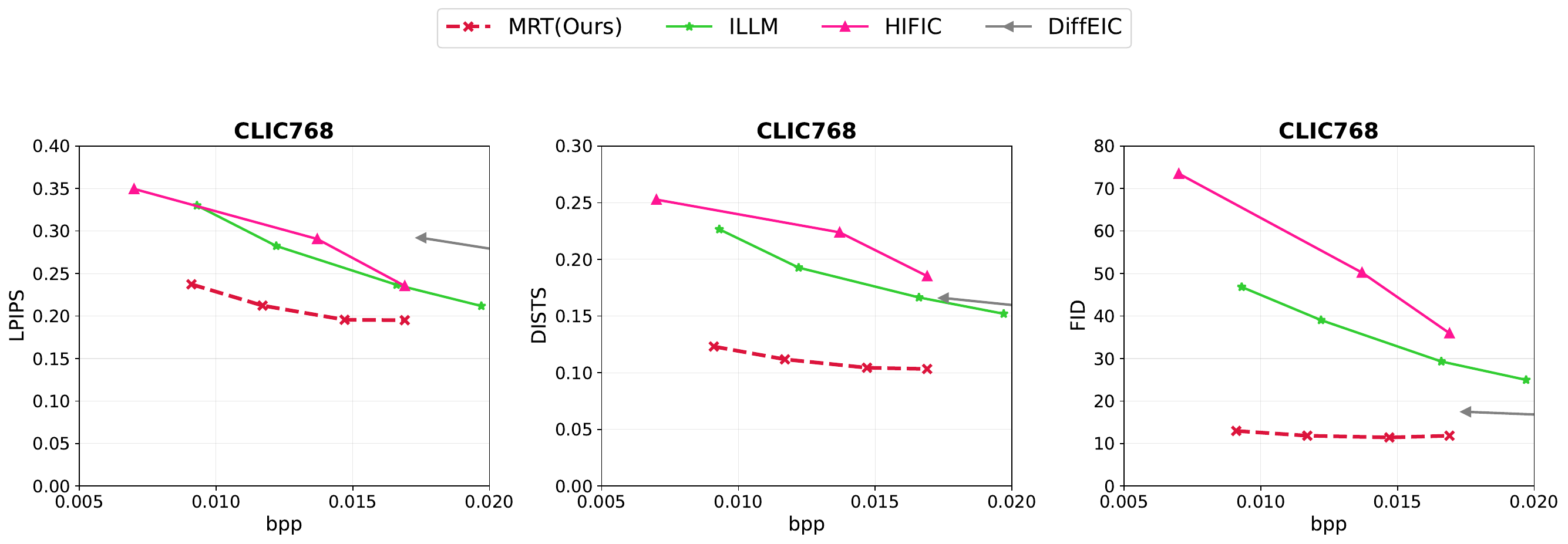} 
    \caption{LPIPS, DISTS, and FID results of different methods on CLIC2020 dataset (768$\times$768 resolution).}
    \label{CLIC768_LPIPS_DISTS_FID}
\end{figure*}

To give a more comprehensive evaluation, we provide the quantitative results of MRT on 
CLIC2020 $768\times768$ test set~\cite{CLIC2020}, as shown in Figure~\ref{CLIC768_LPIPS_DISTS_FID}.

\begin{figure*}[t]
    \centering
    \includegraphics[width=0.9\textwidth]{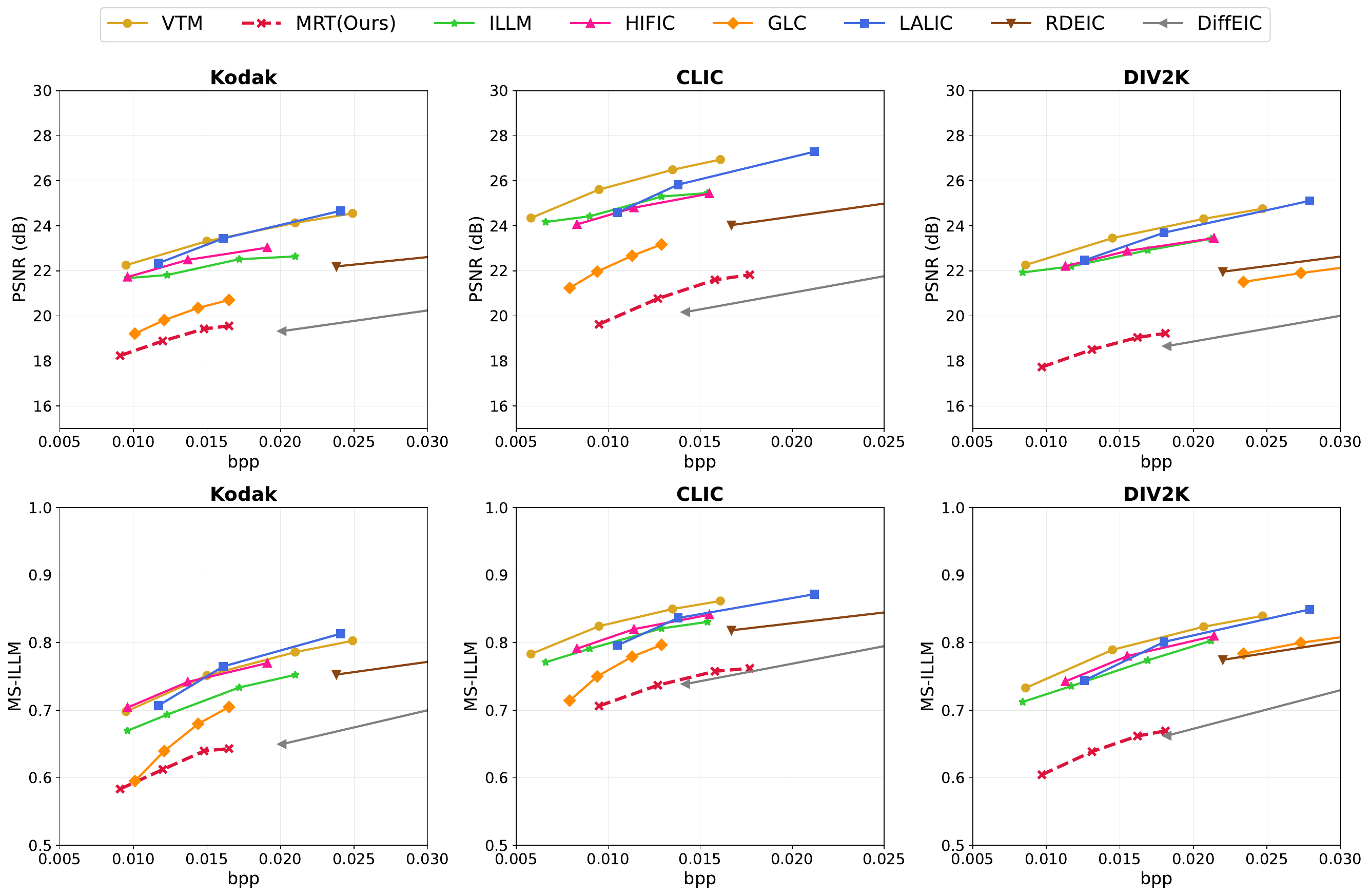} 
    \caption{PSNR and MS-SSIM results of different methods on Kodak, CLIC2020 and DIV2K datasets.}
    \label{PSNR_MS-SSIM}
\end{figure*}

Furthermore, we provide PSNR and MS-SSIM results on Kodak~\cite{kodak}, CLIC2020, and DIV2K~\cite{DIV2K} datasets.
As illustrated in Figure~\ref{PSNR_MS-SSIM}, MRT outperforms the diffusion-based method DiffEIC on all three datasets
in both PSNR and MS-SSIM metrics.
Notably, pixel-level distortion is not well aligned with human perception,
especially at extremely low bitrates~\cite{GLC}. Although MRT does not achieve the best pixel-level rate-distortion performance, it
still exhibits superior visual quality as discussed in the main paper.

\subsection{Qualitative Results}
In this section, we provide additional visual examples across the CLIC2020 test~\cite{CLIC2020}
and MS-COCO 30K~\cite{lin2014microsoft} datasets. For these examples, MRT achieves
pleasing visual quality.

\begin{table}
    \centering
    \begin{tblr}{
      row{1} = {c},
      cell{2}{2} = {c},
      cell{2}{3} = {c},
      cell{3}{2} = {c},
      cell{3}{3} = {c},
      cell{4}{2} = {c},
      cell{4}{3} = {c},
      vline{2} = {-}{},
      hline{1-2,5} = {-}{},
    }
    Kodak & MS-ILLM (2D) & MRT (1D) \\
    MFC ↓ & 0.30 & $\textbf{0.20}$ \\
    cosine ↓ & 0.79 & $\textbf{0.30}$ \\
    $L_2$ distance ↑ & 2.46 & $\textbf{27.69}$ 
    \end{tblr}
    \caption{Redundancy analysis of 2D representations generated by MS-ILLM~\cite{illm} and 1D representations generated by MRT.}
    \label{redundancy}
\end{table}

\begin{table*}
    \centering
    \begin{tblr}{
      row{2} = {c},
      column{10} = {c},
      cell{1}{1} = {r=2}{},
      cell{1}{2} = {c=4}{c},
      cell{1}{6} = {c=4}{c},
      cell{1}{10} = {r=2}{},
      cell{3}{2} = {c},
      cell{3}{3} = {c},
      cell{3}{4} = {c},
      cell{3}{5} = {c},
      cell{3}{6} = {c},
      cell{3}{7} = {c},
      cell{3}{8} = {c},
      cell{3}{9} = {c},
      cell{4}{2} = {c},
      cell{4}{3} = {c},
      cell{4}{4} = {c},
      cell{4}{5} = {c},
      cell{4}{6} = {c},
      cell{4}{7} = {c},
      cell{4}{8} = {c},
      cell{4}{9} = {c},
      cell{5}{2} = {c},
      cell{5}{3} = {c},
      cell{5}{4} = {c},
      cell{5}{5} = {c},
      cell{5}{6} = {c},
      cell{5}{7} = {c},
      cell{5}{8} = {c},
      cell{5}{9} = {c},
      vline{2} = {1-5}{},
      hline{1,3,6} = {-}{},
    }
    Models & Kodak &   &   &   & DIV2K &   &   &   & Params(M) \\
      & Enc.(s) & Dec.(S) & Mem.(G) & BD-DISTS & Enc.(s) & Dec.(S) & Mem.(G) & BD-DISTS &   \\
    MS-ILLM & $0.035$ & $0.030$ & $1.051$ & $0.00\%$ & $0.401$ & $0.361$ & $5.088$ & $0.00\%$ & $181.0$ \\
    DiffEIC & $0.146$ & $3.908$ & $6.690$ & $-34.52\%$ & $0.910$ & $58.875$ & $21.237$ & $40.90\%$ & $1379.50$  \\
    MRT(ours) & $0.164$ & $0.191$ & $4.738$ & $-77.64\%$ & $0.771$ & $1.056$ & $15.730$ & $-76.43\%$ & $906.11$ 
    \end{tblr}
    \caption{Computational complexity analysis of different methods. All tests are conducted on a single NVIDIA A100 GPU.}
    \label{complexity}
\end{table*}

\begin{figure*}[t]
    \centering
    \includegraphics[width=0.9\textwidth]{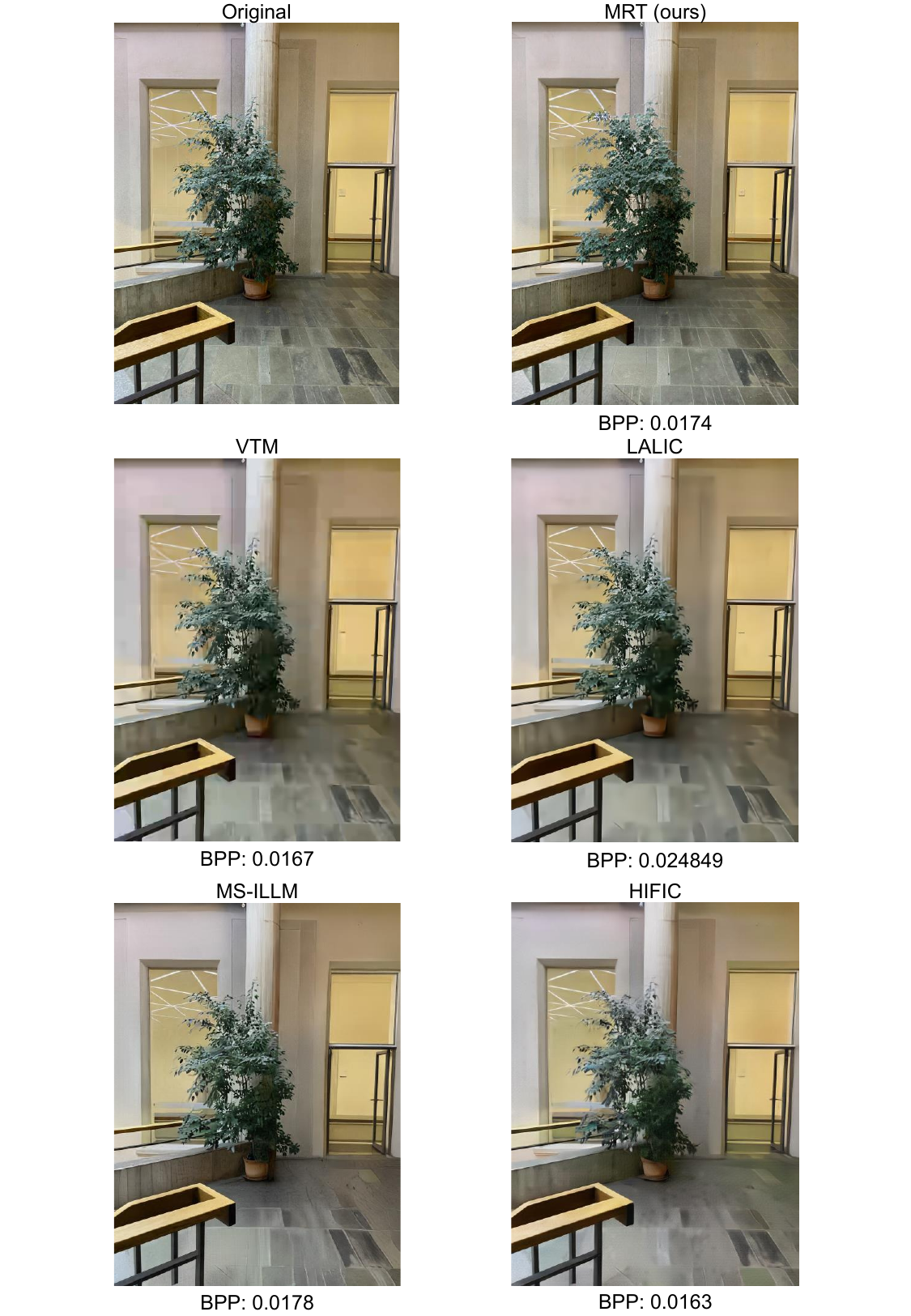} 
    \caption{Qualitative examples on CLIC2020 dataset (full resolution).}
    \label{CLIC}
\end{figure*}

\begin{figure*}[t]
    \centering
    \includegraphics[width=0.9\textwidth]{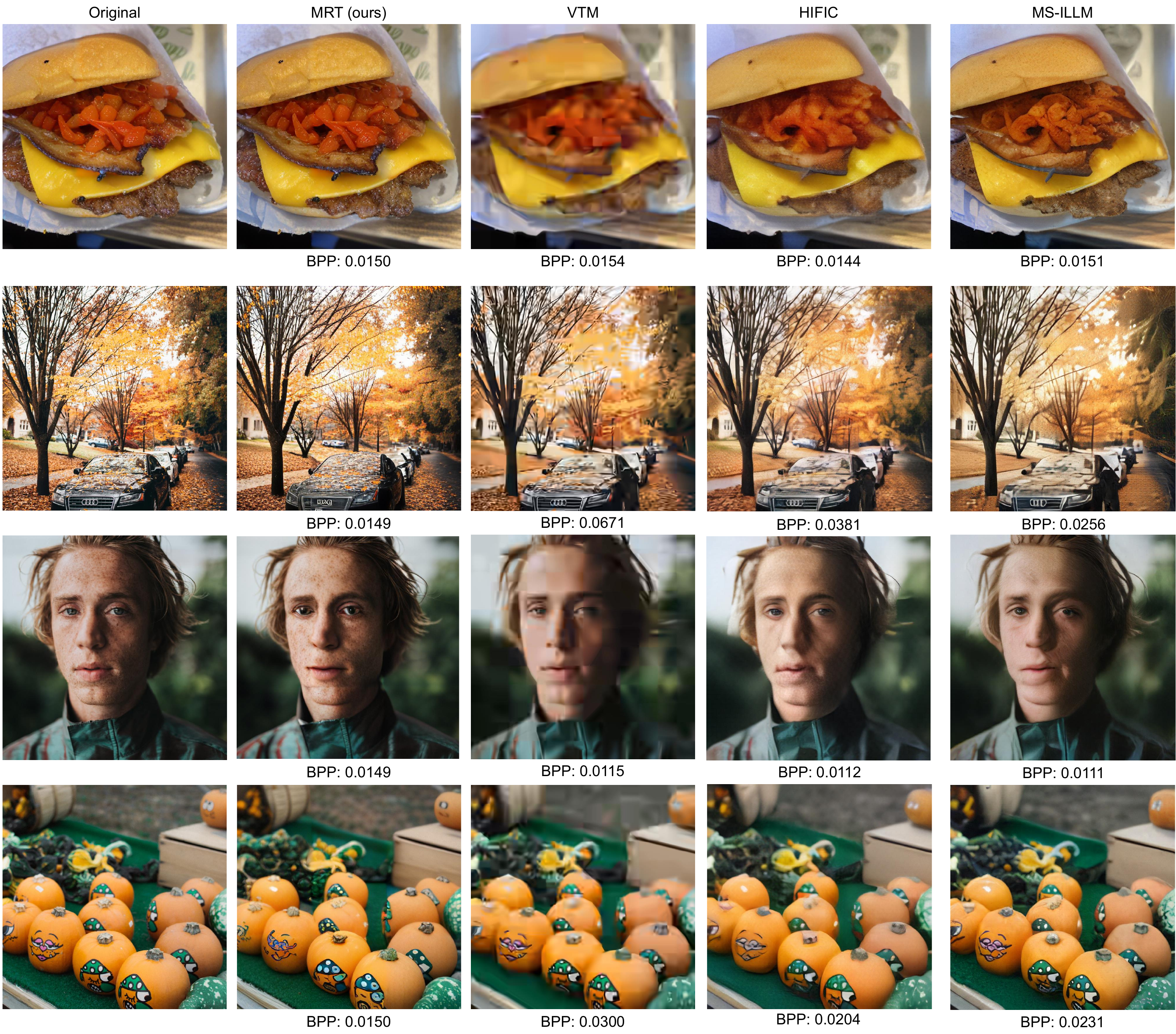} 
    \caption{Qualitative examples on CLIC2020 dataset (768$\times$768 resolution).}
    \label{CLIC768}
\end{figure*}

\begin{figure*}[t]
    \centering
    \includegraphics[width=0.9\textwidth]{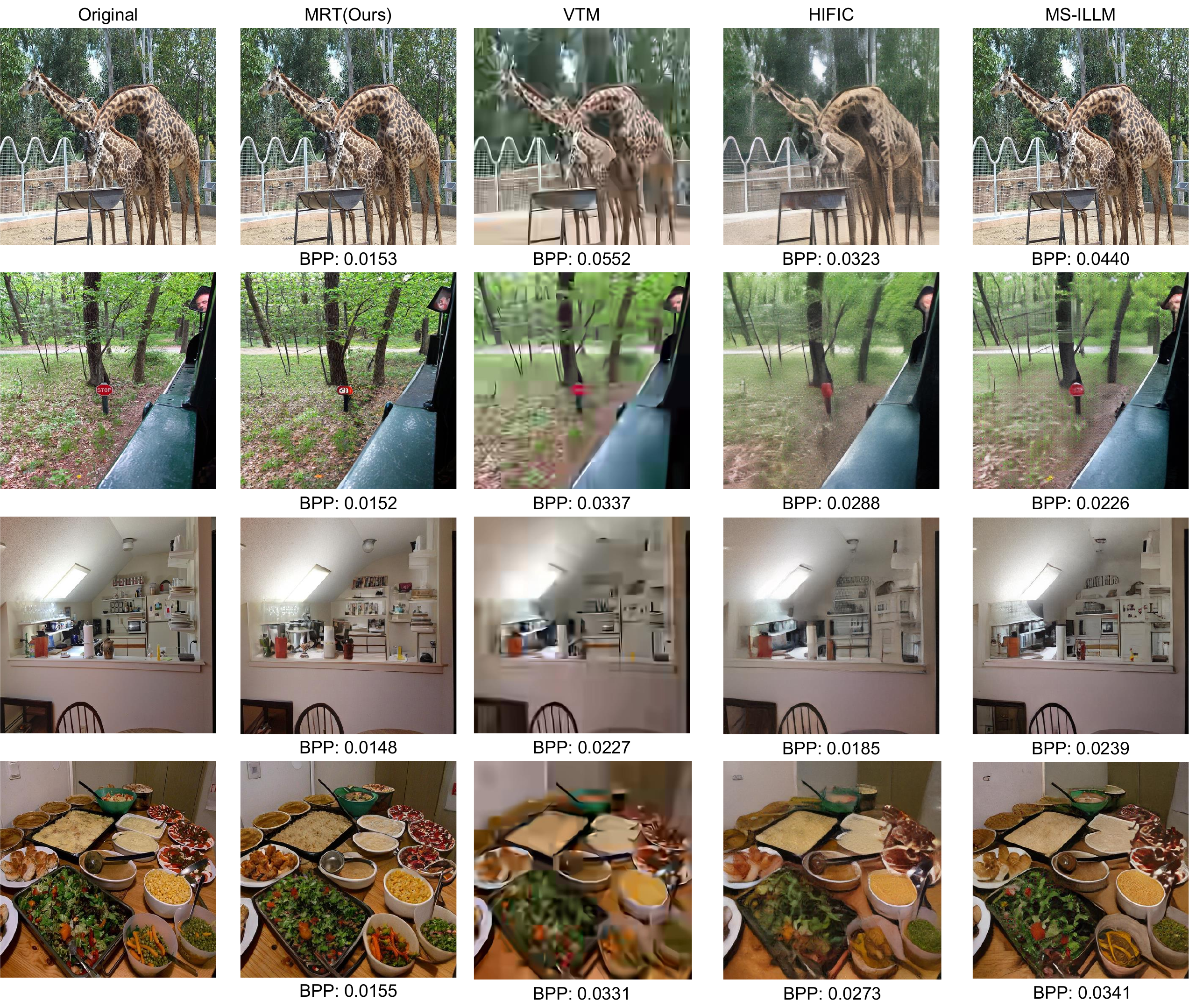} 
    \caption{Qualitative examples on MS-COCO 30K dataset.}
    \label{COCO}
\end{figure*}

\end{document}